\documentclass[a4paper]{article}

\usepackage{INTERSPEECH2020}
\usepackage{times}
\usepackage{latexsym}
\usepackage{adjustbox}
\usepackage{multirow}
\usepackage{arydshln}
\usepackage{amsthm}
\usepackage{todonotes}

\usepackage{url}
\usepackage{graphicx}
\usepackage{caption}
\usepackage{subcaption}
\usepackage{array}
\usepackage{scrextend}
\usepackage{hhline}

\usepackage{hhline}
\usepackage{array,multirow}
\usepackage{arydshln}

\title{To BERT or Not To BERT: Comparing Speech and Language-based Approaches for Alzheimer’s Disease Detection}

\name{Aparna Balagopalan$^1$, Benjamin Eyre$^1$, Frank Rudzicz$^{2,3}$, Jekaterina Novikova$^1$}
\address{
  $^1$Winterlight Labs Inc, Toronto, Canada\\
  $^2$Department of Computer Science / Vector Institute for Artificial Intelligence, Toronto, Canada\\
  $^3$Li Ka Shing Knowledge Institute, St Michael’s Hospital, Toronto, Canada
 }
\email{aparna@winterlightlabs.com, benjamin@winterlightlabs.com, frank@cs.toronto.edu, jekaterina@winterlightlabs.com}

\begin{document}

\maketitle

\begin{abstract}

Research related to automatically detecting Alzheimer's disease (AD) is important, given the high prevalence of AD and the high cost of traditional methods. Since AD significantly affects the content and acoustics of spontaneous speech, natural language processing and machine learning provide promising techniques for reliably detecting AD. We compare and contrast the performance of two such approaches for AD detection on the recent ADReSS challenge dataset~\cite{luz2020alzheimer}: 1) using domain knowledge-based hand-crafted features that capture linguistic and acoustic phenomena, and 2) fine-tuning Bidirectional  Encoder Representations from Transformer (BERT)-based sequence classification models.  We also compare multiple feature-based regression models for a neuropsychological score task in the challenge. We observe that fine-tuned BERT models, given the relative importance of linguistics in cognitive impairment detection, outperform feature-based approaches on the AD detection task.
\end{abstract}

\section{Introduction}

Alzheimer's disease (AD) is a progressive neurodegenerative disease that causes problems with memory, thinking, and behaviour. 
AD affects over 40 million people worldwide with high costs of acute and long-term care~\cite{prince2016world}. Current forms of  diagnosis are both time consuming and expensive \cite{prabhakaran2018analysis}, which might explain why almost half of those living with AD do not receive a timely diagnosis~\cite{jammeh2018machine}. 

Studies have shown that valuable clinical information indicative of cognition can be obtained from spontaneous speech elicited using pictures~\cite{goodglass2001bdae}. Several studies have used speech analysis, natural language processing (NLP), and ML to distinguish between healthy and cognitively impaired speech of participants in picture description datasets~\cite{fraser2016linguistic,zhu2018semi}. These serve as quick, objective, and non-invasive assessments of an individual's cognitive status. However, although ML methods for automatic AD-detection using such speech datasets achieve high classification performance (between 82\%-93\% accuracy)~\cite{fraser2016linguistic, noorian2017importance, karlekar2018detecting}, the field still lacks publicly-available, balanced, and standardised benchmark datasets. The ongoing ADReSS challenge~\cite{luz2020alzheimer} 
provides an age/sex-matched balanced speech dataset, which consists of speech from AD and non-AD participants describing a picture. The challenge consists of two key tasks: 1) Speech classification task: classifying speech as AD  or non-AD.\label{challenge_task:1} 2) Neuropsychological score regression task: predicting Mini-Mental State Examination (MMSE)~\cite{cockrell2002mini} scores from speech.\label{challenge_task:2}
 
In this work, we develop ML models to detect AD from speech using picture description data of the demographically-matched ADReSS Challenge speech dataset~\cite{luz2020alzheimer}, and  
compare the following training regimes and input representations to detect AD:
\begin{enumerate}
    \item \textbf{Using domain knowledge}: with this approach, we extract linguistic features from transcripts of speech, and acoustic features from corresponding audio files for binary AD vs non-AD classification and MMSE score regression. The features extracted are informed by previous clinical and ML research in the space of cognitive impairment detection~\cite{fraser2016linguistic}.
    \item \textbf{Using transfer learning}: with this approach, we fine-tune pre-trained BERT
    ~\cite{devlin2019bert} text classification models at transcript-level. BERT achieved state-of-the-art results on a wide variety of NLP tasks when fine-tuned~\cite{devlin2019bert}. Our motivation is to benchmark a similar training procedure on transcripts from a pathological speech dataset, and evaluate the effectiveness of high-level language representations from BERT in detecting AD.
\end{enumerate}
In this paper, we evaluate performance of these two methods on both the ADReSS train dataset, and on the unseen test set. 
We find that fine-tuned BERT-based text sequence classification models achieve the highest AD detection accuracy 
with an accuracy 
of $83.3$\% on the test set. 
With the feature-based models, the highest accuracy of $81.3$\% is achieved by the SVM with RBF kernel model.
The lowest root mean squared error obtained for the MMSE prediction task is $4.56$, with a feature-based L$2$ regularized linear regression model.

The main contributions of our paper are as follows:
\begin{itemize}
    \item We employ a domain knowledge-based approach and compare a number of AD detection and MMSE regression models with an extensive list of pre-defined linguistic and acoustic features as input representations from speech (Section~\ref{sec:experiments} and \ref{sec:results}).  
    \item We employ a transfer learning-based approach and benchmark fine-tuned BERT models for the AD vs non-AD classification task (Section~\ref{sec:experiments} and \ref{sec:results}).  
    \item We contrast the  performance of the two approaches on the classification task, and discuss the reasons for existing differences (Section~\ref{sec:discussion}).  
\end{itemize}
\section{Background}
\subsection{Domain Knowledge-based Approach}
Previous work has focused on automatic AD detection from speech using acoustic features (such as zero-crossing rate, Mel-frequency cepstral coefficients) and linguistic features (such as proportions of various part-of-speech (POS) tags~\cite{orimaye2015learning, fraser2016linguistic, noorian2017importance}) from speech transcripts. Fraser \emph{et al.}~\cite{fraser2016linguistic} extracted $370$ linguistic and acoustic features from picture descriptions in the DementiaBank dataset, and obtained an AD detection accuracy of $82$\% at transcript-level. More recent studies showed the addition of normative data helped increase accuracy up to $93$\% ~\cite{noorian2017importance,balagopalan2018effect} .


Yancheva \emph{et al}.~\cite{yancheva2015using} showed ML models are capable of predicting the MMSE scores from features of speech elicited via picture descriptions, with mean absolute error of $2.91$-$3.83$.

Detecting AD or predicting scores like MMSE with pre-engineered features of speech and thereby infusing domain knowledge into the task has several advantages, such as more interpretable model decisions and potentially lower resource requirement (when paired with conventional ML models). However, there are also a few disadvantages, e.g. a time consuming process of feature engineering, and a risk of missing highly relevant features.

\subsection{Transfer Learning-based Approach}
In the recent years, transfer learning in the form of pre-trained language models has become ubiquitous in NLP~\cite{young2018recent} and has contributed to the state-of-the-art on a wide range of tasks. One of the most popular transfer learning models is 
BERT~\cite{devlin2019bert}, which builds on Transformer networks~\cite{vaswani2017attention} to pre-train bidirectional representations of text by conditioning on both left  and  right contexts  jointly  in  all  layers. 

BERT uses powerful attention mechanisms to encode global dependencies between the input and output. 
This allows it to achieve state-of-the-art results on a suite of benchmarks~\cite{devlin2019bert}. 
Fine-tuning BERT for a few epochs can potentially attain good performance even on small datasets. However, such models are not directly interpretable, unlike feature-based ones.  

\section{Dataset}
\label{sec:dataset}

We use the ADReSS Challenge dataset~\cite{luz2020alzheimer}, which consists of 156 speech samples and associated transcripts from non-AD ($N$=78) and AD ($N$=78) English-speaking participants. Speech is elicited from participants through the Cookie Theft picture from the Boston Diagnostic Aphasia exam~\cite{goodglass2001bdae}.
In contrast to other speech datasets for AD detection such as DementiaBank's English Pitt Corpus~\cite{becker1994natural}, the ADReSS challenge dataset is matched for age and gender (Table~\ref{tab:ds_compare}). 
The speech dataset is divided into standard train and test sets. 
MMSE~\cite{cockrell2002mini} scores are available for all but one of the participants in the train set. 

\begin{table}[t]
\begin{center}
\caption{Basic characteristics of the patients in each group in the ADReSS challenge dataset are more balanced in comparison to DementiaBank.}

\begin{adjustbox}{max width=0.8\linewidth}
\begin{tabular}{l|cl|c|c}
 \textbf{Dataset} & &  & \multicolumn{2}{c}{\textbf{Class}} \\
 & &  & \textbf{AD} & \textbf{Non-AD}\\ \hhline{=|==|=|=}
\multirow{2}{*}{ADReSS} & \multirow{2}{*}{Train} & Male & 24 & 24 \\

& & Female & 30 &  30 \\\cline{1-5}
\multirow{2}{*}{ADReSS} & \multirow{2}{*}{Test} & Male & 11 &  11 \\
 & & Female & 13 & 13  \\
\cline{1-5}
\multirow{2}{*}{DementiaBank~\cite{becker1994natural}} & \multirow{2}{*}{-} & Male & 125 &  83 \\
  & & Female & 197 & 146  \\
\cline{1-5}
\end{tabular}
\end{adjustbox}
\label{tab:ds_compare}
\end{center}
\end{table}

\section{Feature Extraction}
\label{sec:feature_extract}
The speech transcripts in the dataset are manually transcribed as per the CHAT protocol~\cite{macwhinney2014childes}, and include speech segments from both the participant and an investigator. We only use the portion of the transcripts corresponding to the participant. Additionally, 
we combine all participant speech segments corresponding to a single picture description for extracting acoustic features. 

We extract 509 manually-engineered features from transcripts and associated audio files (see Appendix~\ref{app:feat_list} for a list of all features).
These features are identified as indicators of cognitive impairment in previous literature, and hence encode domain knowledge. All of them are divided into 3 categories:
\begin{enumerate}
    \item \textbf{Lexico-syntactic features (297):}  Frequencies of various production rules from the constituency parsing tree of the transcripts~\cite{chae2009predicting}, speech-graph based features~\cite{mota2012speech}, lexical norm-based features (e.g. average sentiment valence of all words in a transcript, average imageability of all words in a transcript~\cite{warriner2013norms}), features indicative of lexical richness. We also extract syntactic features~\cite{ai2010web} such as the proportion of various POS-tags, and similarity between consecutive utterances.
    \item \textbf{Acoustic features (187):} Mel-frequency cepstral coefficients (MFCCs), fundamental frequency, statistics related to zero-crossing rate, as well as proportion of various pauses~\cite{davis2009examining} (for example, filled and unfilled pauses, ratio of a number of pauses to a number of words etc.)
    \item \textbf{Semantic features based on picture description content (25):} Proportions of various information content units used in the picture, identified as being relevant to memory impairment in prior literature~\cite{croisile1996comparative}.
\end{enumerate}

\section{Experiments}
\label{sec:experiments}
\subsection{AD vs non-AD Classification}

\subsubsection{Training Regimes}
We benchmark the following training regimes for classification: classifying features extracted at transcript-level and a BERT model fine-tuned on transcripts. 

\textbf{Domain knowledge-based approach:}
We classify lexicosyntactic, semantic, and acoustic features extracted at transcript-level with four conventional ML models (SVM, neural network (NN), random forest (RF), na\"ive Bayes (NB)\footnote{https://scikit-learn.org/stable/}.

\textit{Hyperparameter tuning:}
We optimize each model to the best possible hyper-parameter setting using grid-search 10-fold cross-validation (CV). 
We perform feature selection by choosing top-k number of features, based on ANOVA F-value between label/features. The number of features is jointly optimized with the classification model parameters (see Appendix~\ref{app:hyperparam} for a full list of parameters). 

\textbf{Transfer learning-based approach:}
In order to leverage the language information encoded by BERT
~\cite{devlin2019bert}, we use pre-trained model weights to initialize our classification model. We add a classification layer mapping representations from the final BERT layer to binary class labels~\cite{wolf2019huggingface} for the AD vs non-AD classification task. The model is fine-tuned on training data 
with 10-fold CV.
Adam optimizer~\cite{kingma2014adam} and linear scheduling for the learning rate~\cite{paszke2019pytorch} are used. 

\textit{Hyperparameter tuning:}
We optimize the number of epochs to 10 by varying it from 1 to 12 during CV. 
Learning rate and other optimization parameters (scheduling, optimizers etc.) are set based on prior work on fine-tuning BERT ~\cite{devlin2019bert,wolf2019huggingface}.

\begin{table*}[h]
\centering
{
\caption{Feature differentiation analysis results for the most important features, based on ADReSS train set. $\mu_{AD}$ and $\mu_{non-AD}$ show the means of the 13 significantly different features at p$<$9e-5 (after Bonferroni correction) for the AD and non-AD group respectively. 
We also show Spearman correlation between MMSE score and features, and regression weights of the features associated with the five greatest and five lowest regression weights from our regression experiments. * next to correlation indicates significance at p$<$9e-5.
}
\begin{adjustbox}{max height=2cm}

\begin{tabular}{lccccc}
\label{tab:feature-stats}
\textbf{Feature} & \textbf{Feature type} & $\mu_{AD}$ & $\mu_{non-AD}$ & \textbf{Correlation} & \textbf{Weight} \\
\hhline{======}
Average cosine distance between utterances & Semantic  & $0.91$ &  $0.94$& - & -\\
Fraction of pairs of utterances below a similarity threshold (0.5)  & Semantic &  $0.03$& $0.01$ & - & - \\
Average cosine distance between 300-dimensional word2vec~\cite{mikolov2013distributed} utterances and picture content units & Semantic (content units)  & $0.46$& $0.38$ & -0.54* & -1.01\\
Distinct content units mentioned: total content units & Semantic (content units)  & $0.27$ & $0.45$ & 0.63* & 1.78\\
Distinct action content units mentioned: total content units & Semantic (content units)  &$0.15$ & $0.30$ & 0.49* & 1.04\\
Distinct object content units mentioned: total content units  & Semantic (content units) &$0.28$ &  $0.47$ & 0.59* & 1.72\\
Average cosine distance between 50-dimensional GloVe utterances and picture content units & Semantic content units) & - & - & -0.42* & -0.03 \\
Average word length (in letters) & Lexico-syntactic &$3.57$ & $3.78$ & 0.45* & 1.07 \\
Proportion of pronouns   & Lexico-syntactic &$0.09$ & $0.06$ & - & -\\
Ratio (pronouns):(pronouns$+$nouns)   & Lexico-syntactic &$0.35$ & $0.23$ & - & -\\
Proportion of personal pronouns & Lexico-syntactic &$0.09$ & $0.06$ & - & -\\
Proportion of RB adverbs  & Lexico-syntactic &$0.06$ & $0.04$ & -0.41* & -0.41\\
Proportion of ADVP\_$->$\_RB amongst all rules    & Lexico-syntactic &$0.02$ & $0.01$ & -0.37 & -0.74\\
Proportion of non-dictionary words    & Lexico-syntactic & $0.11$ & $0.08$ & - & -\\
Proportion of gerund verbs & Lexico-syntactic  & - & - & 0.37 & 1.08 \\
Proportion of words in adverb category & Lexico-syntactic  & - & - & -0.4* & -0.49 \\

\end{tabular}
\end{adjustbox}
}
\end{table*}

\subsubsection{Evaluation}
\textbf{Cross-validation on ADReSS train set:}
We use two CV strategies in our work -- leave-one-subject-out CV (LOSO CV) and 10-fold CV. We report evaluation metrics with LOSO CV for all models except fine-tuned BERT for direct comparison with challenge baseline. 
 Due to computational constraints of GPU memory, we are unable to perform LOSO CV for the BERT model. Hence, we perform 10-fold CV to compare feature-based classification models with fine-tuned BERT. Values of performance metrics for each model are averaged across three runs of 10-fold CV with different random seeds.
\newline \textbf{Predictions on ADReSS test set:} We generate three predictions with different seeds from each hyperparameter-optimized classifier trained on the complete train set, and then produce a majority prediction to avoid overfitting.
We report performance on the challenge test set, 
as obtained from the challenge organizers 
(see Appendix~\ref{app:test_performance} for more details).

We evaluate task performance primarily using accuracy scores, since all train/test sets are known to be balanced. We also report precision, recall, specificity and F1 with respect to the positive class (AD).


\subsection{MMSE Score Regression}
\subsubsection{Training Regimes}
\textbf{Domain knowledge-based approach:} For this task, we benchmark two kinds of regression models, linear and ridge
, using pre-engineered features as input. MMSE scores are always within the range of 0-30, and so predictions are clipped to a range between 0 and 30.

\textit{Hyperparameter tuning:}
Each model's performance is optimized using hyperparameters selected from grid-search LOSO CV. 
We perform feature selection by choosing top-k number of features, based on an F-Score computed from the correlation of each feature with MMSE score. 
The number of features is optimized for all models. For ridge regression, the number of features is jointly optimized with the coefficient for L2 regularization, $\alpha$.

\subsubsection{Evaluation} We report root mean squared error (RMSE) and mean absolute error (MAE) for the predictions produced by each of the models on the training set with LOSO CV. 
In addition, we include the RMSE for two models' predictions on the ADReSS test set. Hyperparameters for these models were selected based on performance in grid-search 10-fold cross validation on the training set, motivated by the thought that 10-fold CV better demonstrates how well a model will generalize to the test set.

\section{Results}
\label{sec:results}
\subsection{AD vs non-AD Classification}
In Table~\ref{tab:bert_class}, the classification performance with all the models evaluated on the train set via 10-fold CV is displayed. We observe that BERT outperforms all domain knowledge-based ML models with respect to all metrics. SVM is the best-performing domain knowledge-based model. However, accuracy of the fine-tuned BERT model is not significantly higher than that of the SVM classifier based on an Kruskal-Wallis H-test ($H = 0.4838$, $p>0.05$).

We also report the performance of all our classification models with LOSO CV. 
Each of our classification models outperform the challenge baselines by a large margin (+$30$\% accuracy for the best performing model).
It is important to note that feature selection results in accuracy increase of about $13$\% for the SVM classifier. 

Performance results on the unseen, held-out challenge test set are shown in Table~\ref{tab:test_class} 
and follow the trend of the cross-validated performance in terms of accuracy, with BERT outperforming the best feature-based classification model SVM.

\begin{table}[h]
\caption{10-fold CV results averaged across 3 runs with different random seeds on the ADReSS train set.  \label{tab:bert_class} Accuracy for BERT is higher, but not significantly so from SVM ($H=0.4838, p>0.05$ Kruskal-Wallis H test). Bold indicates the best result.}

{
\begin{adjustbox}{max width=0.95\linewidth}

\begin{tabular}[t]{lcccccc}
Model & \#Features & Accuracy &Precision &Recall&Specificity&F1\\
\hhline{=======}
SVM &   10 & $0.796$ &$0.81$ & $0.78$ & $0.82$ & $0.79$ \\
NN &   10 & $0.762$ & $0.77$ & $0.75$ & $0.77$ & $0.76$\\
RF &   50 & $0.738$ & $0.73$ & $0.76$ & $0.72$ & $0.74$ \\
NB &   80 & $0.750$ & $0.76$ & $0.74$ & $0.76$ & $0.75$\\
BERT &  - & $\mathbf{0.818}$  & $\mathbf{0.84}$& $\mathbf{0.79}$ & $\mathbf{0.85}$ & $\mathbf{0.81}$ \\
\end{tabular}
\end{adjustbox}
}
\end{table}

\begin{table}[h!]
\caption{LOSO-CV results averaged across 3 runs with different random seeds on the ADReSS train set. Accuracy for SVM is significantly higher than NN ($H=4.50, p=0.034$ Kruskal-Wallis H test). Bold indicates the best result. \label{tab:cross_class}
}
{
\begin{adjustbox}{max width=0.95\linewidth}

\begin{tabular}[t]{lcccccc}
Model & \#Features & Accuracy &Precision &Recall&Specificity&F1\\
\hhline{=======}
 Baseline~\cite{luz2020alzheimer} & -&$0.574$&$0.57$&$0.52$&-&$0.54$\\
SVM &  509 & $0.741$ &$0.75$ & $0.72$ & $0.76$  & $0.74$ \\
SVM &  10& $\mathbf{0.870}$ &$\mathbf{0.90}$ & $\mathbf{0.83}$ & $\mathbf{0.91}$& $\mathbf{0.87}$ \\
NN &  10& $0.836$ &$0.86$ & $0.81$ & $0.86$ & $0.83$ \\
RF &  50& $0.778$ &$0.79$ & $0.77$ & $0.79$ & $0.78$\\
NB & 80& $0.787$ &$0.80$ & $0.76$ & $0.82$ & $0.78$ \\
\end{tabular}
\end{adjustbox}
}
\end{table}

\begin{table}[h]
\caption{Results on unseen, held-out ADReSS test set \label{tab:test_class}.We present test results in same format as the baseline paper~\cite{luz2020alzheimer}. Bold indicates the best result.}

{
\begin{adjustbox}{max width=0.95\linewidth}

\begin{tabular}[t]{lccccccc}
Model & \#Features&Class & Accuracy &Precision &Recall&Specificity&F1\\
\hhline{========}
\multirow{2}{*}{Baseline~\cite{luz2020alzheimer}} & \multirow{2}{*}{-}& 
non-AD &\multirow{2}{*}{$0.625$} & $0.67$ & $0.50$ & -&$0.57$\\
& &  AD & & $0.60$ & $0.75$ & - & $0.67$\\
\cline{1-8}
\multirow{2}{*}{SVM} & \multirow{2}{*}{$10$}& 
non-AD &\multirow{2}{*}{$0.813$} & $0.83$ & $0.79$ & - &$0.81$\\
& &  AD & & $0.80$ & $0.83$ & - & $0.82$\\
\cline{1-8}
\multirow{2}{*}{NN} & \multirow{2}{*}{$10$}& 
non-AD &\multirow{2}{*}{$0.771$} & $0.78$ & $0.75$ & - &$0.77$\\
& &  AD & & $0.76$ & $0.79$ & - & $0.78$\\
\cline{1-8}
\multirow{2}{*}{RF} & \multirow{2}{*}{$50$}& 
non-AD &\multirow{2}{*}{$0.750$} & $0.71$ & $\mathbf{0.83}$ & - &$0.77$\\
& &  AD & & $0.80$ & $0.67$ & - & $0.73$\\
\cline{1-8}
\multirow{2}{*}{NB} & \multirow{2}{*}{$80$}& 
non-AD &\multirow{2}{*}{$0.729$} & $0.69$ & $\mathbf{0.83}$ & - &$ 0.75$\\
& &  AD & & $0.79$ & $0.63$ & - & $0.70$\\
\cline{1-8}
\multirow{2}{*}{BERT} & \multirow{2}{*}{-}& 
non-AD &\multirow{2}{*}{$\mathbf{0.833}$} & $\mathbf{0.86}$ & $0.79$ & - &$ \mathbf{0.83}$\\
& &  AD & & $0.81$ & $\mathbf{0.88}$ & - & $\mathbf{0.84}$\\

\end{tabular}
\end{adjustbox}
}
\end{table}

\subsection{MMSE Score Regression}
Performance of regression models evaluated on both train and test sets is shown in Table~\ref{tab:cross_reg}. 
Ridge regression with 25 features selected attains the lowest RMSE of 4.56 during LOSO-CV on the training set, a decrease of 2.7 from the challenge baseline. The results show that feature selection is impactful for performance and helps achieve a decrease of up to 1.5  RMSE points (and up to 0.86 of MAE) for a ridge regressor. Furthermore, a ridge regressor is able to achieve an RMSE of 4.56 on the ADReSS test set, a decrease of 1.6 from the baseline.

\begin{table}[h]
\caption{LOSO-CV MMSE regression results on the ADReSS train and test sets. Bold indicates the best result. \label{tab:cross_reg}
}
\centering
{
\begin{adjustbox}{max width=0.7\linewidth}


\begin{tabular}{lcc|cc|c}
Model & \multicolumn{1}{l}{\#Features} & \multicolumn{1}{l|}{$\alpha$} & \multicolumn{1}{l}{RMSE} & \multicolumn{1}{l|}{MAE} & \multicolumn{1}{l}{RMSE} \\
 & \multicolumn{1}{l}{} & \multicolumn{1}{l|}{} & \multicolumn{2}{c|}{Train set} & Test set \\ \cline{2-2}
 \hhline{===|==|=}
Baseline~\cite{luz2020alzheimer} & - & - & 7.28 &  & 6.14 \\
\cline{1-6}
LR & 15 & - & 5.37 & 4.18 & 4.94 \\
LR & 20 & - & 4.94 & 3.72 & - \\
Ridge & 509 & 12 & 6.06 & 4.36 & - \\
Ridge & 35 & 12 & 4.87 & 3.79 & \textbf{4.56} \\
Ridge & 25 & 10 & \textbf{4.56} & \textbf{3.50} & - \\
\end{tabular}
\end{adjustbox}
}
\end{table}

\section{Discussion}
\label{sec:discussion}
\subsection{Feature Differentiation Analysis}
\label{sec:feature_analysis}

We extract a large number of features to capture a wide range of linguistic and acoustic phenomena, based on a survey of prior literature in automatic cognitive impairment detection~\cite{fraser2016linguistic,yancheva2015using,pou2018learning,zhu2019detecting}.
In order to identify the most differentiating features between AD and non-AD speech, we perform independent $t$-tests between feature means for each class in the ADReSS training set. $87$ features are significantly different between the two groups at $p<0.05$. 79 of these are text-based lexicosyntactic and semantic features, while 8 are acoustic. These 8 acoustic features include the number of long pauses, pause duration, and mean/skewness/variance-statistics of various MFCC coefficients.
However, after Bonferroni correction for multiple testing, we identify that only 13 features are significantly different between AD and non-AD speech at $p<9e-5$, and none of these features are acoustic (Table ~\ref{tab:feature-stats}). This implies that linguistic features are particularly differentiating between the AD/non-AD classes here, which explains why models trained on linguistic features only attain performance well above random chance (see Fig.~\ref{fig:tsne} in Appendix for visualization of class separability).

\subsection{Analysing AD Detection Performance Differences}
Comparing classification performance across all model settings, we observe that BERT outperforms the best domain knowledge-based model in terms of accuracy and all performance metrics with respect to the positive class both on the train set (10-fold CV; though accuracy is not significantly higher) and on the test set (no significance testing possible since only single set of performance scores are available per model; see Appendix~\ref{app:test_performance} for procedure for submitting challenge predictions). 
 Based on feature differentiation analysis
 , we hypothesize that good performance with a text-focused BERT model on this speech classification task is due to the strong utility of linguistic features on this dataset. BERT captures a wide range of linguistic phenomena due to its training methodology, potentially encapsulating most of the important lexico-syntactic and semantic features. It is thus able to use information present in the lexicon, syntax, and semantics of the transcribed speech after fine-tuning~\cite{jawahar2019does}.
 \newline We also see a trend of better performance when increasing the number of folds (see SVM in Table~\ref{tab:cross_class} and Table~\ref{tab:bert_class}) in cross-validation. We postulate that this is due to the small size of the dataset, and hence differences in training set size in each fold ($N_{train}=107$ with LOSO, $N_{train}=98$ with 10-fold CV).


\subsection{Regression Weights}
To assess the relative importance of individual input features for MMSE prediction, we report features with the five highest and five lowest regression weights in  Table~\ref{tab:feature-stats}. Each presented value is the average weight assigned to that feature across each of the LOSO CV folds. We also present the correlation with MMSE score coefficients for those $10$ features, as well as their significance, 
in Table~\ref{tab:feature-stats}. We observe that for each of these highly weighted features, a positive or negative correlation coefficient is accompanied by a positive or negative regression weight, respectively. This demonstrates that these $10$ features are so distinguishing that, even in the presence of other regressors, their relationship with MMSE score remains the same. We also note that all $10$ of these are linguistic features, further demonstrating that linguistic information is particularly distinguishing when it comes to predicting the severity of a patient's AD.

\section{Conclusions}
In this paper, we 
compare two widely used approaches -- explicit features engineering based on domain knowledge, and transfer learning using fine-tuned BERT classification model. Our results 
show that pre-trained models that are fine-tuned for the AD classification task 
are capable of performing well on AD detection, and outperforming hand-crafted feature engineering. 
A direction for future work 
is developing ML models that combine representations from BERT and hand-crafted features~\cite{yu2015combining}. Such feature-fusion approaches could potentially boost performance on the cognitive impairment detection task.

\section{Acknowledgements}
This paper benefited greatly from feedback and review from multiple people. Most notably, Dr. J. Robin (Winterlight), L. Kaufman (Winterlight) and M. Yancheva (Winterlight). F. Rudzicz is supported by a CIFAR Chair in AI.

\clearpage
\bibliographystyle{IEEEtran}
\small
\bibliography{mybib}
\appendix

\clearpage

\section{List of features}
\label{app:feat_list}
List of lexico-syntactic features is in Table~\ref{tab:lex_features}, acoustic features in Table~\ref{tab:acoustic_features} and semantic in Table~\ref{tab:semantic_features}, all with brief descriptions and counts of sub-types.


\begin{table*}[ht]
\centering
\caption{Summary of all lexico-syntactic features extracted. The number of features in each subtype is shown in the second column (titled "\#features").\label{tab:lex_features}
}
{
\begin{adjustbox}{max width=\linewidth}

\begin{tabular}{lcc}
 \textbf{Feature type} & \textbf{\#Features} & \textbf{Brief Description} \\
\hhline{===}
Syntactic Complexity & 36 & L2 Syntactic Complexity Analyzer~\cite{lu2010automatic} features; max/min utterance length, depth of syntactic parse tree\\
\cline{1-3}
Production Rules &104 & Number of times a production type occurs divided by total number of productions \\
\cline{1-3}
Phrasal type ratios & 13 & Proportion, average length and rate of phrase types\\
\cline{1-3}
\multirow{2}{*}{Lexical norm-based}
 & \multirow{2}{*}{12} & Average norms across all words, across nouns only and across verbs only for imageability, \\
& & age of acquisition, familiarity and frequency (commonness)\\
\cline{1-3}
Lexical richness & 6 & Type-token ratios (including moving window); brunet; Honoré’s statistic\\
\cline{1-3}
\multirow{3}{*}{Word category} & \multirow{3}{*}{5} & Proportion of demonstratives (e.g., "this"), function words,
\\ 
& & light verbs and inflected verbs, and propositions (POS tag verb, adjective, adverb,\\ & & conjunction, or preposition)\\
\cline{1-3}
Noun ratio & 3 & Ratios nouns:(nouns+verbs); nouns:verbs; pronouns:(nouns+pronouns)\\
\cline{1-3}
Length measures & 1 & Average word length\\
\cline{1-3}
Universal POS proportions & 18 & Proportions of Spacy univeral POS tags~\cite{honnibal2017spacy}\\
\cline{1-3}
POS tag proportions & 53 & Proportions of Penn Treebank~\cite{marcus1994penn} POS tags\\
\cline{1-3}
Local coherence & 15 & Avg/max/min similarity between word2vec~\cite{mikolov2013distributed} representations of utterances (with different dimensions)\\
\cline{1-3}
Utterance distances & 5 & Fraction of pairs of utterances below a similarity threshold (0.5,0.3,0); avg/min distance\\
\cline{1-3}
Speech-graph features & 13 & Representing words as nodes in a graph and computing density, number of loops etc.\\
\cline{1-3}
Utterance cohesion & 1 & Number of switches in verb tense across utterances divided by total number of utterances\\
\cline{1-3}
Rate &  2 & Ratios -- number of words: duration of audio; number of syllables: duration of speech, \\
\cline{1-3}
Invalid words & 1 & Proportion of words not in the English dictionary\\
\cline{1-3}
Sentiment norm-based & 9 & Average sentiment valence, arousal and dominance across all words, noun and verbs\\
\end{tabular}
\end{adjustbox}
}
\end{table*}

\begin{table*}[ht]
\centering
\caption{Summary of all acoustic features extracted. The number of features in each subtype is shown in the second column (titled "\#features").\label{tab:acoustic_features}
}
{
\begin{adjustbox}{max width=\linewidth}

\begin{tabular}{lcc}
 \textbf{Feature type} & \textbf{\#Features} & \textbf{Brief Description} \\
\hhline{===}
\multirow{2}{*}{Pauses and fillers} &\multirow{2}{*}{9} &  Total and mean duration of pauses;long and short pause counts;\\
& & pause to word ratio; fillers(um,uh); duration of pauses to word durations\\
\cline{1-3}
Fundamental frequency & 4 & Avg/min/max/median fundamental frequency of audio\\
\cline{1-3}
Duration-related & 2 & Duration of audio and spoken segment of audio \\
\cline{1-3}
Zero-crossing rate & 4 & Avg/variance/skewness/kurtosis of zero-crossing rate\\
\cline{1-3}
Mel-frequency Cepstral Coefficients (MFCC) & 168 & Avg/variance/skewness/kurtosis of 42 MFCC coefficients\\

\end{tabular}
\end{adjustbox}
}
\end{table*}

\begin{table*}[ht]
\centering
\caption{Summary of all semantic features extracted. The number of features in each subtype is shown in the second column (titled "\#features").\label{tab:semantic_features}
}
{
\begin{adjustbox}{max width=\linewidth}

\begin{tabular}{lcc}
 \textbf{Feature type} & \textbf{\#Features} & \textbf{Brief Description} \\
\hhline{===}
Word frequency  & 10 & Proportion of lemmatized words, relating to the Cookie Theft picture content units to total number of content units\\
\cline{1-3}
Global coherence & 15 & Avg/min/max cosine distance between word2vec~\cite{mikolov2013distributed} utterances and picture content units, with varying dimensions of word2vec \\
\end{tabular}
\end{adjustbox}
}
\end{table*}

\section{Hyper-parameter Settings}
\label{app:hyperparam}
Hyper-parameters were tuned using grid search with 10-fold cross validation on the ADReSS challenge `train' set.

The random forest classifier fits 200 decision trees and considers $\sqrt{features}$ when looking for the best split. The minimum number of samples required to split an internal node is 2, and the minimum number of samples required to be at a leaf node is 2. Bootstrap samples are used when building trees. All other parameters are set to the default value.

The Gaussian Naive Bayes classifier is fit with balanced priors and variance smoothing coefficient set to $1e-10$ and all other parameters default in each case..

The SVM is trained with a radial basis function kernel with kernel coefficient($\gamma$) $0.001$, and regularization parameter set to $100$. 

The NN used consists of 2 layers of 10 units each (note we varied both the number of units and number of layers while tuning for the optimal hyperparameter setting). The ReLU activation function is used at each hidden layer. The model is trained using Adam for 200 epochs and with a batch size of number of samples in train set in each fold. All other parameters are default.

\section{t-SNE Visualization}
  
\begin{figure}[h!]
\caption{A t-SNE plot showing class separation. Note we only use the 13 features significantly different between classes (see Table~\ref{tab:feature-stats}) in feature representation for this plot. \label{fig:tsne}}
\includegraphics[width=0.7\linewidth]{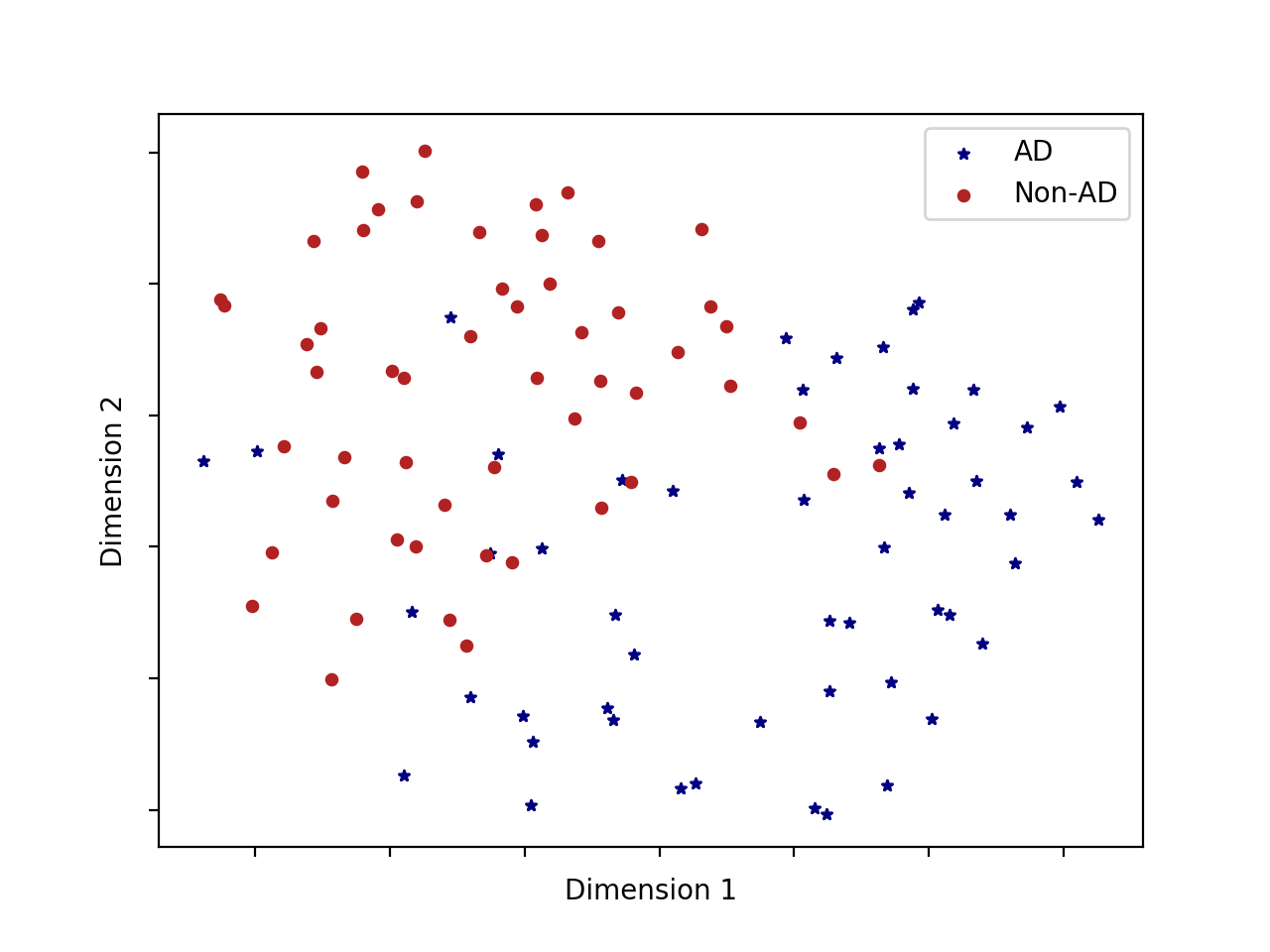}
\centering
\end{figure}

In order to visualize the class-separability of the feature-based representations, we visualize t-SNE~\cite{maaten2008visualizing} plots in Figure~\ref{fig:tsne}. We observe strong class-separation between the two classes, indicating that a non-linear model would be capable of good AD detection performance with these representations.

\section{Test Performance Metrics}
\label{app:test_performance}
The procedure for obtaining performance metrics on the test set was as follows:
\begin{enumerate}
    \item Predictions from up to 5 models are sent to the challenge organizer for each prediction task --  we sent predictions from 5 AD vs non-AD classification models (SVM, NN, RF, NB, BERT) and 5 linear regression models. 
    \item Organizers send performance scores on the test set for each prediction set, which are then reported in Table~\ref{tab:test_class} and Table~\ref{tab:cross_reg}.
\end{enumerate}

\end{document}